\def\year{2020}\relax
\documentclass[letterpaper]{article} 
\usepackage{aaai20}  
\usepackage{times}  
\usepackage{helvet} 
\usepackage{courier}  
\usepackage[hyphens]{url}  
\usepackage{graphicx} 
\usepackage{graphicx}  
\usepackage{latexsym}
\usepackage{multirow}
\usepackage{url}
\usepackage{xcolor}
\usepackage{multicol}
\usepackage{booktabs}
\usepackage{tabularx}
\usepackage{todonotes}
\usepackage{comment} 
\usepackage{enumerate} 
\usepackage{multibib}

\title{Discourse Level Factors for  Sentence Deletion in Text Simplification}

\author{Yang Zhong,\textsuperscript{\rm *1} Chao Jiang,\textsuperscript{\rm 1} Wei Xu,\textsuperscript{\rm 1} Junyi Jessy Li\textsuperscript{\rm 2} \\ \textsuperscript{1} Department of Computer Science and Engineering, The Ohio State University\\
  \textsuperscript{2} Department of Linguistics, The University of Texas at Austin \\
  {\tt \{zhong.536, jiang.1530, xu.1265\}@osu.edu \quad jessy@austin.utexas.edu}\\
}

\date{}

\begin{document}
\maketitle
\begin{abstract}

This paper presents a data-driven study focusing on analyzing and predicting sentence deletion --- a prevalent but understudied phenomenon in document simplification --- on a large English text simplification corpus. We inspect various document and discourse factors associated with sentence deletion, using a new manually annotated sentence alignment corpus we collected. We reveal that professional editors utilize different strategies to meet readability standards of elementary and middle schools. To predict whether a sentence will be deleted during simplification to a certain level, we harness automatically aligned data to train a classification model. Evaluated on our manually annotated data, our best models reached F1 scores of 65.2 and 59.7 for this task at the levels of elementary and middle school, respectively. We find that discourse level factors contribute to the challenging task of predicting sentence deletion for simplification.
{\let\thefootnote\relax\footnote{* Work started as an undergraduate student at UT Austin.}} 
\end{abstract}
\section{Introduction}
Text simplification aims to rewrite an existing document to be accessible to a broader audience (e.g., non-native speakers, children, and individuals with language impairments) while remaining truthful in content. The simplification process involves a variety of operations, including lexical and syntactic transformations, summarization, removal of difficult content, and explicification \cite{surveyOnTS}. 

While recent years saw a bloom in text simplification research \cite{Xu-EtAl:2016:TACL,narayan-gardent-2016-unsupervised,nisioi-etal-2017-exploring,zhang-lapata:2017:EMNLP2017,vu-etal-2018-sentence,sulem-etal-2018-simple,maddela-xu-2018-word,Kriz-et-al:2019:NAACL} thanks to the development of large parallel corpora of original-to-simplified sentence pairs \cite{zhu-etal-2010-monolingual,tacl:Xu}, most of the recent work is conducted at the sentence-level, i.e., transducing each complex sentence to its simplified version.  

As a result, this line of work does not capture document-level phenomena, among which sentence deletion is the most prevalent, as simplified texts tend to be shorter \cite{DBLP:conf/slte/PetersenO07,Drndarevic2012ReducingTC,woodsend2011wikisimple}.

\begin{figure*}[t]
   \centering
   \includegraphics[width=0.85\textwidth]{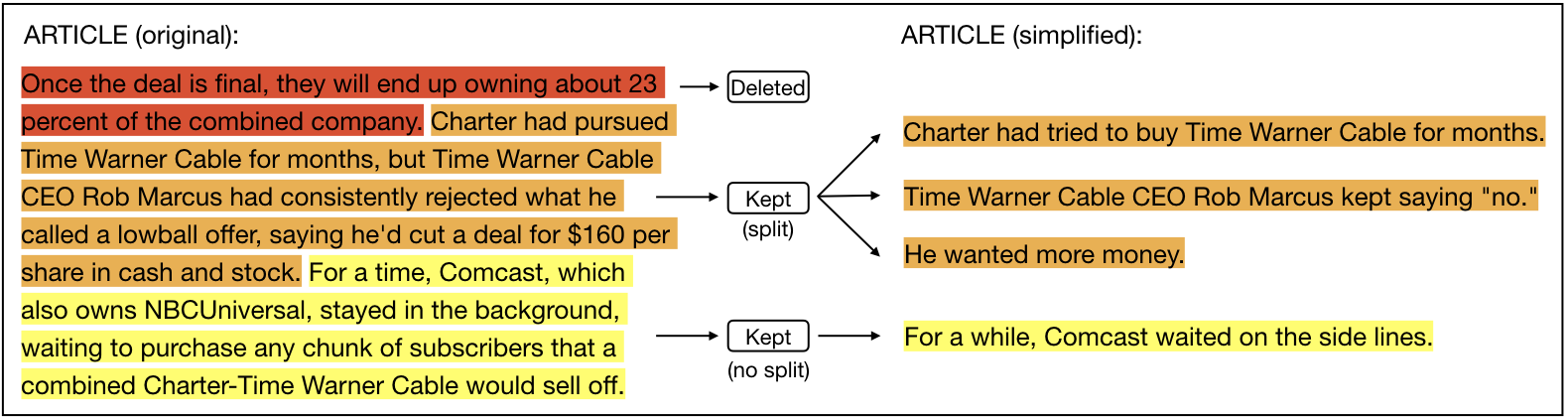}
   \caption{An example paragraph of the original news article (left) and its simplified version by professional editors for elementary school students (right). The arrows signify the sentence alignment between the original and simplified documents. The second sentence in the original paragraph is aligned to three sentences after simplification. }\label{fig:example}
\end{figure*}

This work aims to facilitate better understanding of sentence deletion in document-level text simplification. While prior work analyzed sentence position and  content in a corpus study \cite{DBLP:conf/slte/PetersenO07}, we hypothesize and show that sentence deletion is driven partially by contextual, discourse-level information, in addition to the content within a sentence. 

We utilize the Newsela\footnote{Newsela is an educational tech company that provides reading material for children in school curricula.} corpus~\cite{tacl:Xu} which contains multiple versions of a document rewritten by professional editors. This dataset allows us to compare sentence deletion strategies to meet different readability requirements. Unlike prior work that often uses automatically aligned data for analysis (c.f.\ Section \ref{relatedwork}), 
we manually aligned 50 articles of more than 5,000 sentences across three reading levels to provide a reliable ground truth for the analysis and model evaluation in this paper.\footnote{To request our data, please first obtain access to the Newsela corpus at: https://newsela.com/data/, then contact the authors.} We find that sentence deletion happens very often at rates of 17.2\%--44.8\% across reading levels, indicating that sentence deletion prediction is an important task in text simplification. Several characteristics of the original document, including its length and topic, significantly influence sentence deletion. By analyzing the rhetorical structure \cite{RST} of the original articles, we show that the sentence deletion process is also informed by how a sentence is situated in terms of its connections to neighboring sentences, and its discourse salience within a document. In addition, we reveal that the use of discourse connectives within a sentence also influence whether it will be deleted. 

To predict whether a sentence in the original article will be deleted during simplification, we utilize noisy supervision obtained from 886 automatically aligned articles with a total of 42,264 sentences. Our neural network model learns from both the content of the sentence itself, as well as the discourse level factors we analyzed. Evaluated on our manually annotated data, our best model that utilizes Gaussian-based feature vectorization achieved F1 scores of 65.2 and 59.7 for this task across two different reading levels (elementary and middle school). We show that several of the factors, especially document characteristics, complements sentence content in this challenging task. On the other hand, while our analysis of rhetorical structure revealed interesting insights, encoding them as features does not further improve the model. To the best of our knowledge, this is the first data-driven study that focuses on analyzing discourse-level factors and predicting sentence deletion on a large English text simplification corpus.

\section{Data and Setup}
We use the Newsela text simplification corpus \cite{tacl:Xu} of 936 news articles. Each article set consists of 4 or 5 simplified versions of the original article, ranging from grades 3-12 (corresponding to ages 8-18). We group articles into three reading levels: original (grade 12), middle school (grades 6-8) and elementary school (grades 3-5). We use one version of article from each reading level, and study two document-level transformations: original $\rightarrow$ middle and original $\rightarrow$ elementary.

We conduct analysis and learn to predict if a sentence would be dropped by professional editors when simplifying text to the desired reading levels. To obtain labeled data for analysis and evaluation, we manually align sentences of 50 article sets. The resulting dataset is one of the largest manually annotated datasets for sentence alignment in simplification. Figure \ref{fig:example} shows a 3-sentence paragraph in the original article, aligned to the elementary school version. Sentences in the original article that cannot be mapped to any sentence in a lower reading level are considered deleted. To train models for sentence deletion prediction, we rely on noisy supervision from automatically aligned sentences from the rest of the corpus.

\paragraph{Manual alignment.}
Manual sentence alignment is conducted on 50 sets of the articles (2,281 sentences in the original version), using a combination of crowdsourcing and in-house analysis. The annotation process is designed to be efficient, with rigorous quality control, and includes the following steps:

(1) Align paragraphs between articles by asking in-house annotators to manually verify and correct the automatic alignments generated by the CATS toolkit~\cite{stajner18.630}. 
Automatic methods are much more reliable for aligning paragraphs than sentences, given the longer contexts. We use this step to reduce the number of sentence pairs that need to be annotated. 

(2) Collect human annotations for sentence alignment using Figure Eight,\footnote{\url{https://figure-eight.com}} a crowdsourcing platform. For every possible pair of sentences within the aligned paragraphs, we ask 5 workers to classify it into three categories: meaning equivalent, partly overlapped, or mismatched.\footnote{We provided the following guideline: (i) two sentences are equivalent if they convey the same meaning, though one sentence can be much shorter or simpler than the other sentence; (ii) two sentences partially overlap if they share information in common but some important information differs/is missing; (iii) two sentences are mismatched, otherwise.} The final annotations are aggregated by majority vote. To ensure quality, we embedded a hidden test question in every five questions we asked, and removed workers whose accuracy dropped below 80\% on the test questions. The inter-annotator agreement is 0.807 by Cohen's kappa \cite{artstein-poesio-2008-survey}.

(3) Have four in-house annotators (not authors) to verify the crowdsourced labels and recover a few sentence alignments which are outside the aligned paragraph pairs generated in Step 1.

For this study, we consider a sentence in the original article \textbf{deleted} by the editor during simplification, if there is no corresponding sentence labeled as meaning equivalent or partly overlap in the elementary or middle school levels. For sentences that are shortened or split, we consider them as being \textbf{kept}. For more details about the manually-annotated sentence alignment dataset, please refer to \cite{jiang2020neural}.

\paragraph{Automatic alignment.}

We align sentences between pairs of articles based on the cosine similarity of their vector representations. We use 700-dimensional sentence embeddings pretrained on 16GB English Wikipedia by Sent2Vec \cite{pgj2017unsup}, an unsupervised method that learns to compose sentence embeddings from word vectors along with bigram character vectors. Automatic aligned data includes a total of 42,264 sentences from the original article.

We consider a pair of sentences aligned if their similarity exceeds 0.94 when no sentence splitting is involved, or 0.47 when splitting occurs. These thresholds are calibrated on the manually annotated set of article pairs. Empirically tested on the manually labeled data, this alignment strategy is more accurate (72\% accuracy) than the alignment method used by \cite{tacl:Xu} (67\% accuracy), which is based on Jaccard similarity.

\paragraph{Corpus Statistics.} 
Table \ref{table:1} shows the average and standard deviation of the portion of sentences deleted when an article is being simplified from original to middle or elementary levels.
Notably, the standard deviation of the deletion ratio is high, which reflects the multi-facet nature of sentence deletion in simplification (c.f.\ Section~\ref{sec:anslysis}). Simplifying to the elementary level involves on average 27.6\% more deletion than to the middle school level. We also find that automatic alignment results in a much lower deletion rate, indicating that it over-match sentences. 

\begin{table}[!t]
\centering
\small 
\begin{tabular}{c|c|c}
\toprule
\multirow{2}{*}{} & Middle & Elementary \\
\cmidrule{2-3}
 & Avg. (std) & Avg. (std)  \\
 \midrule
automatic alignment & 0.146 ($\pm$0.158)  & 0.272  ($\pm$0.172) \\
manual alignment & 0.172 ($\pm$0.154) & 0.448 ($\pm$0.161) \\
\bottomrule
\end{tabular}
\caption{Fraction of sentences deleted from the original article to meet middle and elementary school reading standards.}
\label{table:1}
\end{table}


\begin{table}[!t]
\centering
\small
\begin{tabular}{l|rr|rr}
\toprule
 & \multicolumn{2}{c|}{Middle} 
 & \multicolumn{2}{c}{Elementary} \\
\cmidrule{2-5}
 & Corr. &  p-value & Corr. & p-value \\ 
\midrule
\# of sentences & 0.849 & 6.8e-15 & 0.470 & 5.6e-4 \\ 
\# of tokens & 0.845 & 1.2e-14 & 0.487 & 3.3e-4 \\ 
\bottomrule
\end{tabular}
\caption{Pearson correlation between deletion rate and document length, measured by the number of sentences and tokens respectively.}
\label{Col:deletion-ratio}
\end{table}

\section{Analysis of Discourse Level Factors}
\label{sec:anslysis}

We present a series of analyses to study discourse level factors, including document characteristics, rhetorical structure, and discourse relations, that potentially influence sentence deletion during simplification.

\subsection{Document Characteristics}
\paragraph{Document length}\label{doc_length}
We hypothesize that the length of the original article will impact how much content professional editors choose to compress to reach a certain reading level. Table~\ref{Col:deletion-ratio} tabulates the Pearson correlation between the number of sentences and words in the original document versus the number of deleted sentences, on the manually aligned articles. The correlations are significant for both middle and elementary levels, yet with the middle level the correlation values are particularly high. Longer documents indeed have higher percentages of sentences being deleted.

\paragraph{Topics}
Intuitively, different topics could also lead to varied difficulty in reading comprehension. We compare percentages of sentences deleted across different article categories available in the Newsela dataset. We conduct this particular analysis on all articles, including auto aligned ones, as the distribution of topic labels is sparse on the manually aligned subset. Because of the noise, we compare deletion rates {\em relative to the mean}. Shown in Table \ref{table:topic}, topics vary in their deletion rates. {\em Science} articles have significantly lower deletion rates for both middle and elementary levels. Articles about {\em Money} and {\em Law} have significantly higher deletion rates than others. 

\begin{table}[!t]
\centering
\small 
\begin{tabular}{c|c|l|l}
\toprule
 & \# of articles & \multicolumn{1}{c|}{Middle} & %
    \multicolumn{1}{c}{Elementary} \\
 \midrule
Science & 282& $-$ 0.0380$^*$ & $-$ 0.0722$^*$ \\
Health & 92 & $-$ 0.0253 & $-$ 0.0033 \\
Arts & 79& $-$ 0.0200 & $+$ 0.0014 \\
War  & 170& $-$ 0.0192 & $-$ 0.0140 \\
Kids & 179 & $+$ 0.0029 & $+$ 0.0147 \\
Money &160 & $+$ 0.0230$^*$ & $+$ 0.0169 \\
Law & 193 & $+$ 0.0283$^*$ & $+$ 0.0402\\
Sports &95 & $+$ 0.0488 & $+$ 0.0300 \\
\bottomrule
\end{tabular}
\caption{Average difference between deletion rate of each topic and the average. * indicates statistical significance ($p < 0.05$) comparing the distribution of one topic and the mean of all other topics based on the two sample Kolmogorov-Smirnov test \cite{KStest}.}
\label{table:topic}
\end{table}

\subsection{Rhetorical Structure}
\begin{figure}[t]
  \centering
  
   \includegraphics[width=.8\columnwidth]{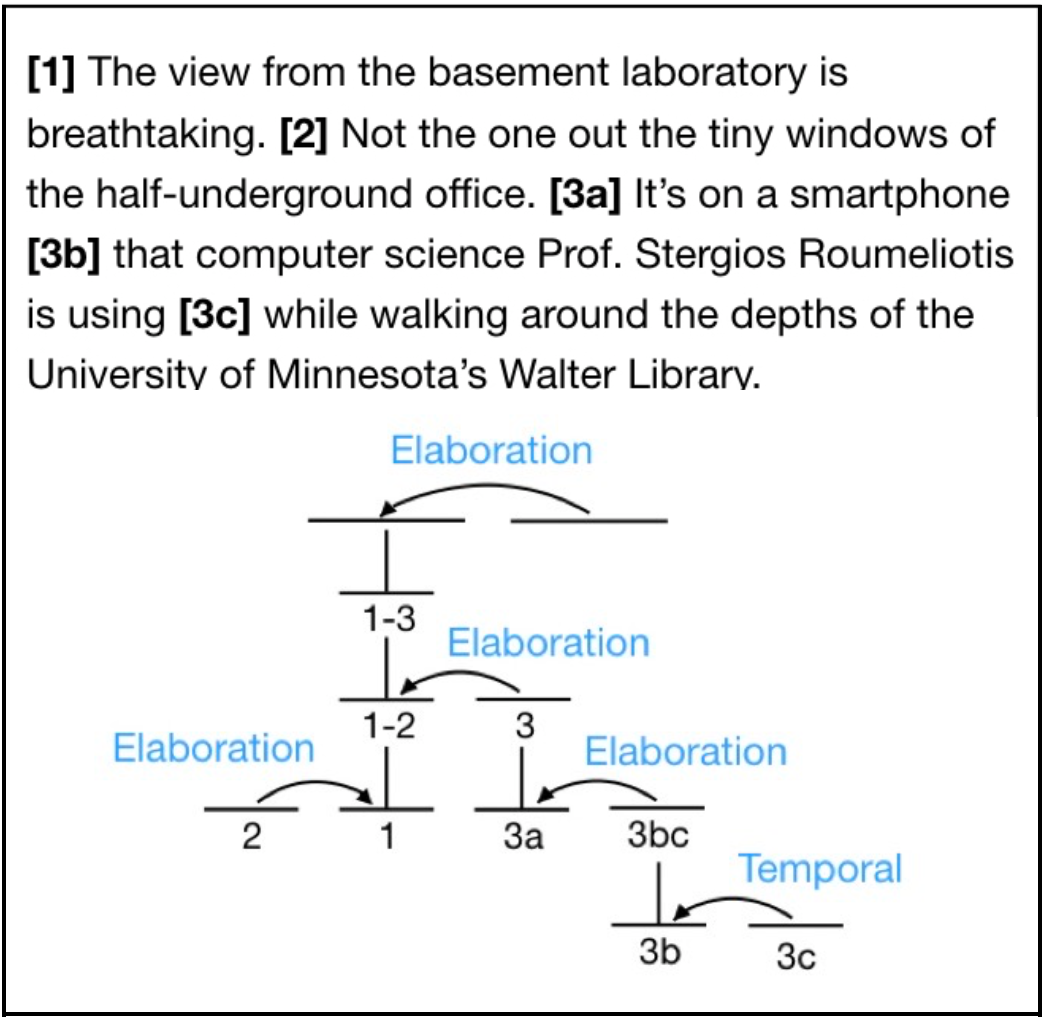}
   \caption{An example RST tree of a segment in an original news article. The arrows represent nucleus (arrow head) and satellite (arrow tail). In the elementary level, [1] is kept and rephrased, [2] is deleted, the third sentence is kept but split into two -- [3a] and [3b] as one sentence, and [3c] as another.}\label{fig:discourse}
\end{figure}

Rhetorical Structure Theory (RST) describes the relations between text spans in a discourse tree, starting from elementary discourse units (roughly, independent clauses). An argument of a relation can be a nucleus (presents more salient information) or a satellite, illustrated in Figure~\ref{fig:discourse}. RST is known to be useful in related applications, including summarization~\cite{MarcuImportance,hirao2013single,durrett2016learning} where information salience plays a central role.

In this section, we focus on how each sentence is situated in the RST tree of the original document, hence we treat each sentence as a discourse unit (that is not necessarily an elementary discourse unit). We use the discourse parser from  \cite{RST_parser} to process each document.

\paragraph{Depth in discourse tree}
RST captures the salience of a sentence with respect to its role in the larger context. In particular, the salience of a unit or sentence does not strictly follow the linear order of appearance in the document, but is indicated by its distance to the highest level of topical organization \cite{cristea1998veins}. Indeed, we found that the relative position of a sentence is not strongly correlated with the depth of the sentence in the discourse tree: the Pearson correlations are 0.064 and -0.088 for kept and deletion conditions at the elementary level, respectively. To this end, we consider the depth of the current sentence in the RST tree of the document (viewing each sentence as a discourse unit). 
 \begin{table}[t]
\centering
\small
\footnotesize
\begin{tabular}{c|cc|cc}
\toprule
 & \multicolumn{2}{c|}{Middle} 
 & \multicolumn{2}{c}{Elementary} \\
\cmidrule{2-5}
&  Kept &  Deleted & Kept &  Deleted \\ 
\midrule
Mean depth &  7.63 & 9.06  & 7.58 &   8.36 \\ 
(std) &  ($\pm$3.05) & ($\pm$3.78)  & ($\pm$3.09) &   ($\pm$3.44) \\ 
\bottomrule
\end{tabular}
\caption{Depth distribution of sentences in RST trees. Deleted sentences are located significantly ($p<0.05$) lower than the ones that are kept, using the Wilcoxon ranked test \cite{wilcoxon1992individual}.}
\label{table:distribution_depth}
\end{table}
The distributions of depth for both deleted and kept sentences are shown in Table \ref{table:distribution_depth}. We observe that sentences that are deleted locate at significantly lower levels in their discourse trees comparing to those that are kept. Since salient sentences tend to locate closer to the root of the discourse tree, this indicates that salience plays some role in the decision whether a sentence should be deleted. 

\paragraph{Nuclearity}

In addition to global salience captured by depth  above, we also look into local information salience. An approximate of the importance of a sentence among its neighboring sentences is the nuclearity information between the relations of the sentence and its neighbors. Therefore, 
we compare how often sentences that are nuclei of their parent relation are deleted vs.\ satellite ones, across each level of simplification. 
While we found that satellite sentences tend to be deleted for the elementary level, the differences are small, and a Chi-Squared test yield no significance ($p=0.2$ for both reading levels).

\subsection{Discourse Relations} \label{discourse}

Discourse relations signal  relationships (e.g., contrast, causal) between clauses and sentences.
These relations capture pragmatic aspects of text and are prominent players in text simplification~\cite{surveyOnTS,siddharthan2003preserving}. Prior work also suggested that sentences in an instantiating role tend to be very detailed~\cite{li-2016-NAACL}, and different relations could indicate different levels of importance for the article as a whole \cite{louis2010discourse}. 
In this work, we look at (a) relations that connect a sentence to the rest of the document, and (b) the usage of explicit discourse connectives within a sentence.

\paragraph{Inter-sentential relations}
We first consider how a sentence is connected with the rest of the document such that its appearance renders the document coherent, especially when information more ``salient'' to this sentence is present and less likely to be deleted. To this end, we study the lowest ancestor relation to which it is attached as a satellite, henceforth the ``governing'' relation. Table \ref{RSTrelation} shows the fraction of sentences belonging to the top 6 governing relations.\footnote{We did not include other relations as their frequencies are less than 0.5\% in the dataset.} Observe that the {\em elaboration} relation is the most frequent relation in the dataset; sentences serving as an elaboration of another sentence are more likely to be removed during simplification (statistically significant for the elementary level). Important sentences that are not satellites to any relation ({\em root}) is significantly less likely to be deleted across both levels. Furthermore, sentences that serve as an {\em explanation} of an existing sentence are less likely to be deleted during simplification (significantly, for the middle school level). 

\begin{table}[!t]
\centering
\small
\begin{tabular}{r|ll|ll}
\toprule
\multirow{3}{*}{} & \multicolumn{2}{c|}{Middle} & %
    \multicolumn{2}{c}{Elementary} \\
    
\cmidrule{2-5}
 & Kept & Deleted & Kept & Deleted  \\

\midrule
Root & 0.084 & 0.057 $\downarrow$ & 0.115 & 0.038 $\downarrow$ \\
Elaboration & 0.793 & 0.816 & 0.752 & 0.840 $\uparrow$\\
Contrast & 0.040 & 0.042 & 0.041 & 0.033 \\
Background & 0.019 & 0.012 & 0.022& 0.021 \\
Evaluation & 0.017 & 0.010 & 0.016 & 0.018 \\
Explanation & 0.019 & 0.011 $\downarrow$ &0.020 & 0.016 \\
\bottomrule
\end{tabular}
\caption{Fraction of discourse relations that govern the sentence's sub-tree. Arrows indicate significance ($p<0.05$) using the Wilcoxon ranked test; $\uparrow$: higher presence among deleted sentences than the kept ones; $\downarrow$: lower.}
\label{RSTrelation}
\end{table}
\paragraph{Discourse connectives}

Discourse connectives are found to have different rates of mental processing in cognitive experiments~\cite{sanders2000role}. 
To identify discourse connectives, we parse each document with the NUS parser \cite{lin2014pdtb} for the Penn Discourse Treebank (PDTB) \cite{Prasad2008ThePD}. 
The fraction of sentences containing a connective is shown in Table~\ref{table:3}. Higher reading levels (original and middle) contain significantly ($p = 1e-48$) more discourse connectives per sentence.

\begin{table}[!t]
\centering
\small
\begin{tabular}{c|c|c|c}
\toprule
\multirow{3}{*}{} & Original & Middle & %
Elementary \\
\cmidrule{2-4}
 & Avg. (std) & Avg. (std) & Avg. (std) \\
 \midrule
 \% of sents & 0.38 ($\pm$0.17) & 0.34 ($\pm$0.16) & 0.22 ($\pm$0.09) \\
  \bottomrule
\end{tabular}
\caption{Fraction of sentences that contain explicit discourse connectives.}
\label{table:3}
\end{table}

We first compare how often discourse connectives appear in deleted vs.\ kept sentences in the original version, and the relation senses they signal: contingency, comparison, expansion, or temporal, following the PDTB taxonomy. We did not conduct analysis on fine-grained levels of the taxonomy due to label sparsity.
Table~\ref{Connective-specific} shows the fraction of sentences containing at least one connective, as well as fraction of sentences containing connectives of a certain sense. 
Deleted sentences are in general significantly more likely to have a connective, a potential signal that the sentence is complex (i.e., has more than one clause). This is especially evident for the elementary level in that this holds for all relations. 
Deleted sentences are significantly less likely to have temporal ones in the middle level. One explanation could be that temporal connectives presuppose the events involved \cite{lascarides-oberlander-1993-temporal}, hence we think they need to be included for the reader to be able to comprehend text as a whole.

We also study the position of these explicit connectives. Specifically, connective positions (start of a sentence vs.\ not) are strong indicators of whether the relation they signal is intra- or inter-sentential~\cite{lin2014pdtb,biran2015pdtb}. 
Table \ref{Col:postion} reveals that when targeting the middle level, sentences with connectives at the beginning (``sent-initial'') are much more likely ($p=9e-6$) to be kept. This could indicate editors being unwilling to delete closely related sentences when the  simplification strategy is less aggressive.

\begin{table}[!t]
\centering
\small
\begin{tabular}{r|ll|ll}
\toprule
\multirow{3}{*}{} & \multicolumn{2}{c|}{Middle} & %
    \multicolumn{2}{c}{Elementary} \\
    
\cmidrule{2-5}
 & Kept & Deleted & Kept & Deleted  \\
 \midrule
 \% of sents & 0.305 & 0.337$\uparrow$ & 0.312 & 0.352$\uparrow$ \\
\midrule
Contingency & 0.077 & 0.079$\uparrow$ & 0.081 &  0.087$\uparrow$ \\

 Comparison & 0.064 & 0.085$\uparrow$ & 0.066  & 0.094$\uparrow$ \\

  Expansion & 0.118 & 0.125 & 0.117  & 0.132$\uparrow$ \\

  Temporal & 0.111 & 0.099$\downarrow$ & 0.098  & 0.107$\uparrow$ \\
\bottomrule
\end{tabular}
\caption{Fraction of sentences that contain the explicit discourse connectives. Arrows indicate significance ($p<0.05$) using the Wilcoxon ranked test; $\uparrow$: higher presence among deleted sentences than the kept ones; $\downarrow$: lower.}
\label{Connective-specific}
\end{table}

\begin{table}[!t]
\centering
\small
\begin{tabular}{r|ll|ll}
\toprule
 & \multicolumn{2}{c|}{Middle} 
 & \multicolumn{2}{c}{Elementary} \\
\cmidrule{2-5}
&  Kept &  Deleted & Kept &  Deleted \\ 
\midrule
Sent-initial & 0.584 & 0.195$\downarrow$ & 0.405 & 0.421 \\ 
Non-initial & 0.740 & 0.147$\downarrow$ &  0.561 & 0.395$\downarrow$ \\ 
\bottomrule
\end{tabular}
\caption{Fraction of sentences with a discourse connective at the start of the sentence or otherwise. $\downarrow$ indicates a significantly ($p<0.05$) lower presence among deleted sentences than the kept ones based on two sample Kolmogorov-Smirnov test.} 
\label{Col:postion}
\end{table}

\section{Predicting Sentence Deletion}
We run our experiments on two tasks, first on building a classification model to see if it can predict whether a sentence should be deleted when simplifying to middle and element level. Second, we perform the feature ablation to determine whether in practice document and discourse signals help under noisy supervision.

\paragraph{Experiment Setup} 
Given a sentence in the original article, we (i) predict whether it will be deleted when simplifying to the middle school level, trained on noisy supervision from automatic alignments; (ii) predict the same for the elementary level. We use 15 of the manually aligned articles as the validation set and the other 35 articles as test set.

\paragraph{Method} We use logistic regression (LR) and feedforward neural networks (FNN) as classifiers,\footnote{We also tried BiLSTM based feature encoding and it gives similar results for the prediction.} and experiment with features from multiple, potentially complementary aspects.  To capture {\bf sentence-level semantics}, we consider the average of GloVe word embeddings \cite{pennington2014glove}. The sparse features (SF) include the relative position of the sentence in the whole article, as well as in the paragraph it resides. Additionally, we include readability scores for the sentence following \cite{scarton-etal-2018-text}\footnote{Flesch Reading Ease, Flesch-Kincaid Grade Level, SMOG Index, Gunning Fog Index, Automated Readability Index, Coleman-Liau Index, Linsear Write Formula and Dale-Chall Readability Score.}.  Leveraging  our corpus analysis (Section~\ref{sec:anslysis}), we incorporate {\bf document-level features}, including the total number of sentences and number of words in the document, as well as the topic of the document. Our {\bf discourse features} include the depth of the current sentence, indicator features for nuclearity and the governing relation of the current sentence in the RST tree, whether there is an explicit connective of one of the four relations we analyzed, and the position of the connective. 
We also use the {\bf position} of the sentence, as  sentences appearing later in an article are more likely to be dropped \cite{DBLP:conf/slte/PetersenO07}.

To improve the prediction performance, we adopted a smooth binning approach \cite{maddela-xu-2018-word} and project each of the sparse features, which are either binary or numerical, into a $k$-dimentional vector representation by applying $k$ Gaussian radial basis functions. 

\paragraph{Implementation Details} 

We use Pytorch to implement the neural network model. To get a sentence representation, we take the average word embeddings as input and stack two hidden layers with ReLU activation, and a single-node linear output layer if only embeddings are used for classification. To combine the learned embedding features and sparse features, we concatenate the output of second hidden layer of the embedding network with the binned sparse features, and feed them into a multi-layer feedforward network with two hidden layers and one last sigmoid layer for classification. We use 300-dimensional GloVe embeddings and a total of 35 sparse features as inputs, and half of the input size nodes in each hidden layer. The training objective is to minimize the binary cross entropy loss between the logits and the true binary labels. We use Adam \cite{adam} for optimization and also apply a dropout of 0.5 to prevent overfitting. We set the learning rate to 1e-5 and 2e-5 for experiments in Tables \ref{font-table2} and \ref{font-table} respectively. We set the batch size to 64. We followed \cite{maddela-xu-2018-word} and set the number of bins $k$ to 10 and the adjustable fraction $\gamma$ to 0.2 for the Gaussian feature vectorization layer. We implemented the logistic regression classifier using Scikit-learn \cite{scikit}. Since our data heavily skews towards keeping all sentences, we use downsampling to balance the two classes.

\begin{table}[!t]
\begin{center}
\small
\begin{tabular}{l|l|l|l}
\toprule \bf Model (Elementary) & \bf Precision & \bf Recall \ & \bf F1 \\  \midrule
Random & 42.1 & 48.8 & 45.2 \\
\midrule
LR Embedding & 58.1 & 58.0 & 58.1 \\
FNN Embedding & 59.2 & 68.0 & 63.2 \\ 
\midrule
LR All Sparse Features & 69.6 & 45.9 & 55.3
 \\
LR All SF binning &  59.9 & 65.7 & 62.7
\\
FNN All Sparse Features & 72.3 & 47.7 & 57.4 \\
FNN All SF binning & 70.2 & 54.0 & 61.0 \\
\midrule
LR Embed \& Sparse Feature & 70.4 & 43.1 & 53.4 \\
LR Embed \& SF binning & 62.2 & 64.0 & 63.1 \\
FNN Embed \& SF binning & 65.9 & 64.5 & 65.2 \\

\bottomrule
\end{tabular}
\end{center}
\caption{\label{font-table2} Performance of predicting sentence deletions for elementary school level simplification. }
\end{table}

\begin{table}[!t]
\small
\begin{center}
\begin{tabular}{l|l|l|l}
\toprule \bf Model (Middle) & \bf Precision \ & \bf Recall \ & \bf F1 \\  \midrule
Random & 21.1 & 49.0 &  29.5 \\
\midrule
LR Embedding & 31.4 & 51.5 & 39.0 \\
FNN Embedding & 34.2 & 61.7 & 44.0 \\
\midrule
LR All Sparse Features & 56.0 & 58.9 & 57.4
 \\
LR All SF binning &  42.4 & 80.1 & 55.4 \\
FNN All Sparse Features & 55.9 & 58.6 & 57.2 \\
FNN All SF binning & 55.9 & 63.8 & 59.6 \\
\midrule
LR Embed \& Sparse Feature & 55.9 & 58.6 & 57.2 \\
LR Embed \& SF binning &41.7 & 75.9 & 53.9 \\
FNN Embed \& SF binning & 56.4 & 63.6 & 59.7\\

\bottomrule
\end{tabular}
\end{center}
\caption{\label{font-table} Performance of predicting sentence deletions for middle school level simplification. }
\end{table}

\begin{table}[!t]
\begin{center}
\small
\begin{tabular}{l|l|l|l}
\toprule \bf Model (Elementary) & \bf Precision & \bf Recall  & \bf F1 \\  \midrule
LR Embed \& SF binning &  62.2 & 64.0 & 63.1 \\
-- Discourse &  62.1 & 63.7 & 62.9 \\
-- Document & 60.7 & 59.2 & 60.0$\downarrow$ \\
-- Position &  58.5 & 62.7 & 60.5$\downarrow$\\
\toprule 
\bf Model (Middle) & \bf Precision  & \bf Recall & \bf F1 \\  \midrule
LR Embed \& SF binning & 41.7 & 75.9 & 53.9 \\
-- Discourse & 41.7 & 75.9 & 53.9 \\
-- Document & 36.0 & 64.1 & 46.1$\downarrow$ \\
-- Position & 39.1 & 78.0 & 52.1$\downarrow$ \\
\bottomrule
\end{tabular}
\end{center}
\caption{\label{font-table3} Feature ablation analysis for predicting sentence deletion by removing one feature category at a time. $\downarrow$ indicates significant drop of performance ($p<0.05$) compared to using all features based on the bootstrapping test \cite{KirkpatrickBK12}. }
\end{table}
\paragraph{Results}
The experimental results are shown in Tables \ref{font-table2} and  \ref{font-table}. As baseline, we consider randomly removing sentences according to deletion rates in the training set. We then look at using the semantic content of the sentences, captured by GLoVe embeddings, and/or using the sparse features. In general, we find this a challenging task. Predicting sentence deletion at the middle level is more difficult than for elementary, as fewer sentences are deleted (c.f.~Table~\ref{table:1}). Comparing the uses of features, we find that middle level deletion and elementary level deletion depend on different features. As shown in Table \ref{font-table2} and Table \ref{font-table}, models using only sentence embedding have a higher performance on element level deletion prediction, yet embeddings are not that informative compared to sparse features in middle level. We also tried the two neighbor sentences' embedding as a bonus feature, but find that they have a negative effects over the middle level deletion task and barely no improvement over element level. This reflects the difficulty in middle level deletion prediction, as professional editors might not use specific strategies to pick up deletion candidates based on their sentence semantics alone. On the other hand, using sparse features only for both levels gives comparable results to the best model that utilizes both categories of features. We also find that the Gaussian binning methods proposed by \cite{maddela-xu-2018-word} significantly help the model to make use of the sparse features, which are initially a small number of discrete features.

For feature ablation, we perform the experiments over the Logistic Regression model since neural models can be heavily influenced by hyper-parameters and random initialization \cite{yang2019critically}. For both levels, document characteristics matter more than position, especially in the middle level task. One reason could be that middle level simplification is based more on content,  while editors tend to shorten the whole texts in elementary level simplification by dropping sentences near the end of an article. 
Overall, the RST and discourse relation features do not help too much, possibly because these features tend to have much lower triggering rates than others, e.g., not every sentence has explicit discourse features as shown in Table \ref{Connective-specific}. Another reason could be the noise introduced during automatic alignment, for example, potential noise in partial match, that could render signals from discourse relations uninformative during training.

\section{Related Work}\label{relatedwork}
Most existing work on text simplification focuses on word/phrase-level \cite{yatskar-etal-2010-sake,P11-2087,specia2012,glavavs-vstajner:2015:ACL-IJCNLP,paetzold2017a,maddela-xu-2018-word} or sentence-level simplifications \cite{zhu-etal-2010-monolingual,Xu-EtAl:2016:TACL,stajner-nisioi-2018-detailed,dong-etal-2019-editnts}.  Only a few projects conducted corpus analyses and automatic prediction on sentence deletion during document-level simplification, including the pioneer work by \citeauthor{DBLP:conf/slte/PetersenO07}~\shortcite{DBLP:conf/slte/PetersenO07}. They analyzed a corpus from Literacyworks (unfortunately, inaccessible by other researchers), and reported the prediction on which sentences will be dropped is ``little better than always choosing the majority class (not dropped)'' using a decision tree based classifier.

In contrast, we study the Newsela corpus which has been widely used among researchers since its release \cite{tacl:Xu}, as it offers a sizable collection of news articles written by professional editors at five different readability levels. It exhibits more significant sentence dropping and discourse reorganization phenomena. We present an in-depth analysis, focusing on various discourse-level factors that are important to understand for developing document-level automatic simplification systems, very different from prior studies of Newsela \cite{tacl:Xu,scarton-etal-2018-text} that focused on vocabulary usage and sentence readability. Other related works include \citeauthor{StajnerEtAl-13}~\shortcite{StajnerEtAl-13}'s on Spanish, \citeauthor{gasperin2009learning}~\shortcite{gasperin2009learning}'s on Brazilian Portuguese, and \citeauthor{Gonzalez-Dios2018}~\shortcite{Gonzalez-Dios2018}'s on Basque. 

More importantly, nearly all the existing studies on sentence deletion and splitting in simplification are based on automatically aligned sentence pairs, without manually labeled ground truth to gauge the reliability of the findings. This is largely due to the scarcity and cost of manually labeled sentence alignment data. In this paper, we present an efficient crowdsourcing methodology and the first manually annotated, high-quality sentence alignment corpus for simplification. To the best of our knowledge, the most comparable dataset is that created by \citeauthor{hwang-etal-2015-aligning}~\shortcite{hwang-etal-2015-aligning} using Wikipedia data, which is inherently noisy as shown by \citeauthor{tacl:Xu}~\shortcite{tacl:Xu}, due to the lack of quality control and strict editing guideline in creating the Simple English Wikipedia. 

\section{Conclusion}
This paper presents a parallel text simplification corpus with manually aligned sentences across multiple reading levels from the Newsela dataset. Our corpus analysis show that discourse-level factors are important when editors drop sentences as they simplify. We further show that document characteristic features help in predicting whether a sentence will be deleted during simplification, a challenging task given the low deletion rate when simplifying to the middle school level. To the best of our knowledge, this is the first data-driven study that focuses on analyzing discourse factors and predicting sentence deletion on a large English text simplification corpus. We hope this work will spur more future research on automatic document simplification.

\section*{Acknowledgments}
We thank NVIDIA and Texas Advanced Computing Center at UT Austin for providing GPU computing resources, the anonymous reviewers and Mounica Maddela for their valuable comments. We also thank Sarah Flanagan, Bohan Zhang, Raleigh Potluri, and Alex Wing for their help with data annotation. This research was partly supported by the NSF under grants IIS-1822754, IIS-1850153, a Figure-Eight AI for Everyone Award and a Criteo Faculty Research Award to Wei Xu. The views and conclusions contained in this publication are those of the authors and should not be interpreted as representing official policies or endorsements of the U.S. Government. 
\bibliographystyle{aaai.bst}
\bibliography{aaai20}

\end{document}